\documentclass[10pt,journal,compsoc,twoside]{IEEEtran}
\ifCLASSOPTIONcompsoc
  \usepackage[nocompress]{cite}
\else
  \usepackage{cite}
\fi

\ifCLASSINFOpdf
   \usepackage[pdftex]{graphicx}
   \graphicspath{{../figures_maintext/}}
   \DeclareGraphicsExtensions{.pdf,.jpg,.eps}
\else
   \usepackage[dvips]{graphicx}
   \graphicspath{{../figures/}}
   \DeclareGraphicsExtensions{.eps}
\fi

\usepackage[dvipsnames]{xcolor}
\usepackage{amsmath}
\usepackage{array}
\usepackage{float}
\usepackage{url}
\usepackage{amsfonts}
\usepackage{algorithmic}
\usepackage{algorithm}
\usepackage[caption=false,font=normalsize,labelfont=sf,textfont=sf]{subfig}
\usepackage{textcomp}
\usepackage{stfloats}
\usepackage{url}
\usepackage{verbatim}
\usepackage{graphicx}
\usepackage{cite}
\usepackage{booktabs}
\usepackage{multirow}
\usepackage{arydshln}
\usepackage{multicol}
\usepackage{xr-hyper}
\usepackage{hyperref}
\usepackage{caption}
\usepackage{threeparttable} 
\usepackage{svg} 
\makeatletter
\newcommand*{\addFileDependency}[1]{
  \typeout{(#1)}
  \@addtofilelist{#1}
  \IfFileExists{#1}{}{\typeout{No file #1.}}
}
\makeatother

\newcommand*{\myexternaldocument}[1]{%
    \externaldocument{#1}%
    \addFileDependency{#1.tex}%
    \addFileDependency{#1.aux}%
}
\myexternaldocument{supplemental}

\newcolumntype{o}{D{.}{.}{1}} 
\newcolumntype{t}{D{.}{.}{3}} 
\newcolumntype{f}{D{.}{.}{4}} 

\makeatletter
\newcolumntype{B}[3]{>{\boldmath\DC@{#1}{#2}{#3}}c<{\DC@end}}
\makeatother

\begin{document}
\bstctlcite{BSTcontrol}
\title{Fairness-enhancing mixed effects deep learning improves fairness on in- and out-of-distribution clustered (non-iid) data}

\author{
        Son~N.~Nguyen,
        Adam~J.~Wang, and
        Albert~A.~Montillo\\
\IEEEcompsocitemizethanks{
    \IEEEcompsocthanksitem Son N. Nguyen and Adam J. Wang contributed equally to this work
    \IEEEcompsocthanksitem Son N. Nguyen and Albert A. Montillo are with Lyda Hill Department
    of Bioinformatics, University of Texas Southwestern Medical Center, Dallas,
    TX, 75390 USA. E-mail: son.nguyen@utsouthwestern.edu, albert.montillo@utsouthwestern.edu\protect
    \IEEEcompsocthanksitem Albert A. Montillo is jointly with the Department of Biomedical Engineering, University of Texas Southwestern Medical Center, Dallas,
    TX, 75390 USA. E-mail: albert.montillo@utsouthwestern.edu
    \IEEEcompsocthanksitem Adam J. Wang is with John A. Paulson School of Engineering and Applied Sciences, Harvard University, Cambridge, MA, 02138 USA.
    \makeatletter
    E-mail: adamwang@college.harvard.edu\protect
}
}

\ifCLASSOPTIONpeerreview
    \markboth{}%
    {Fair MEDL for clustered data}
\else
    \markboth{}%
    {Nguyen, Wang, \& Montillo: Fairness-enhancing mixed effects deep learning}
\fi


\IEEEtitleabstractindextext{%
\begin{abstract}
Traditional deep learning (DL) models have two ubiquitous limitations. First, they assume training samples are independent and identically distributed (i.i.d), an assumption often violated in real-world datasets where samples have additional correlation due to repeat measurements (e.g., on the same participants in a longitudinal study or cells from the same sequencer). This leads to performance degradation, limited generalization, and covariate confounding, which induces Type I and Type II errors. Second, DL models typically prioritize overall accuracy, favoring accuracy on the majority while sacrificing performance for underrepresented subpopulations, leading to unfair, biased models. This is critical to remediate, particularly in models which influence decisions regarding loan approvals and healthcare. To address these issues, we propose the Fair Mixed Effects Deep Learning (Fair MEDL) framework. This framework quantifies cluster-invariant fixed effects (FE) and cluster-specific random effects (RE) through: 1) a cluster adversary for learning invariant FE, 2) a Bayesian neural network for RE, and 3) a mixing function combining FE and RE for final predictions. Fairness is enhanced through architectural and loss function changes introduced by an adversarial debiasing network. We formally define and demonstrate improved fairness across three metrics: equalized odds, demographic parity, and counterfactual fairness, for both classification and regression tasks. Our method also identifies and de-weights confounded covariates, mitigating Type I and II errors. The framework is comprehensively evaluated across three datasets spanning two industries, including finance and healthcare. The Fair MEDL framework improves fairness by 86.4\% for \textit{Age}, 64.9\% for \textit{Race}, 57.8\% for \textit{Sex}, and 36.2\% for \textit{Marital status}, while maintaining robust predictive performance.
\end{abstract}

}
\maketitle
\IEEEdisplaynontitleabstractindextext
\IEEEpeerreviewmaketitle

\IEEEraisesectionheading{\section{Introduction}}
\IEEEPARstart{D}{eep} learning (DL) has become an integral tool in many applications in domains such as finance \cite{shi2022}, healthcare \cite{nguyen2020preoperative, polat2024}, and computer vision \cite{krizhevsky2012imagenet, nguyen2021balanced}. However, the full potential of DL models in such domains is hindered by inherent limitations. Specifically, DL models assume that data samples are independent and identically distributed (\textit{i.i.d}), yet natural datasets frequently violate this assumption. Subsets of samples are correlated due to repeat measures (e.g., multiple samples from the same hospital site, geographic region, or occupation, or through the longitudinal tracking of participants) \cite{connor2008, kozora2008}. This correlation causes sample clustering which can yield poor model prediction performance, confounded analyses, and spurious associations. Failure to address the clustering can result in unreliable models which do not generalize well to out-of-distribution data \cite{goodfellow2016}.

Standard deep learning model training maximizes performance on the majority group within a training set, while sacrificing performance for minority groups \cite{barocas2023}. This is especially harmful in domains like finance and healthcare because: 1) an unfair model can produce inaccurate predictions impacting life-altering decisions, 2) unfairness towards a particular group can lead to those individuals receiving poorer care or inadequate access to loans, and 3) an unfair model perpetuates inequities present in society.  Navarro et al.  outlined such fairness issues in healthcare, highlighting the compromises often made for minority groups \cite{navarro2021risk}. Despite these inequities, an integrated DL framework handling sample clustering \textit{and} enhancing fairness simultaneously has remained unsolved. This work introduces a robust fairness-enhancing mixed effects deep learning (MEDL) framework that yields all of the performance benefits of mixed effects deep learning (e.g., \cite{nguyen2023ARMED}) from addressing clustering, while simultaneously enhancing fairness for mixed and fixed effects predictions. This MEDL framework improves prediction accuracy on clustered data, increases model interpretability by separately inferring cluster-invariant fixed effects (FE) and cluster-dependent mixed effects (ME) predictions, and reduces biases against subpopulations. 
\subsection{Related work}
Several MEDL methods have been introduced as potential solutions to the aforementioned clustering, including MeNets \cite{xiong2019} and LMMNN \cite{simchoni2021}, as well as ARMED \cite{nguyen2023ARMED}, with ARMED comparing favorably across multiple problem domains. Meanwhile, the causes of unfairness in machine learning (ML) \cite{chouldechova2020} can be broadly categorized into those that stem from biases in the data and those that arise from biases in the ML algorithm. This work focuses on alleviating algorithmic biases. Methods for addressing such algorithmic unfairness are subcategorized into: pre-process, in-process, and post-process methods \cite{caton2024fairness}. Our framework uses an \textit{in-process} approach because it imposes the required accuracy-to-fairness trade-off directly in the objective function \cite{woodworth2017}. Pre-process methods can harm the interpretability of the results and leave high uncertainty regarding accuracy at the end of the process \cite{pessach2022review}, and post-process methods require the final decision-maker to possess the sensitive group membership information, which may not be available \cite{woodworth2017}. Among in-process methods, \cite{yang2023adversarial} used adversarial debiasing with a traditional DL multilayer perceptron. They attained an improvement in fairness while classifying COVID-19 severity (high vs. low); however, the authors did not assess the statistical significance of the improvement. \cite{beutel2019correlationRE} introduced an alternative to adversarial debiasing, called absolute correlation loss, which acts as a regularization term for the loss function and can significantly improve the equity of the false positive rate across subpopulations. Neither of these works, however, combined fairness enhancement with a MEDL model to address the bias due to sample clustering. The profound lack of ML research at the intersection of mixed effects modeling and fairness-enhancement motivates our work.

\subsection{Contributions}
This work makes the following contributions: \textbf{1.} We introduce a comprehensive, general framework that simultaneously addresses sample clustering and enhances fairness for subpopulations, as defined by one or more fairness sensitive variables (e.g., \textit{Race}, \textit{Sex}, \textit{Age}), setting the stage for fairer and more reliable machine learning results. \textbf{2.} We rigorously test our method using a diverse set of real-world datasets from finance and healthcare, covering both classification and regression tasks, while measuring fairness with three metrics: equalized odds, demographic parity, and counterfactual fairness. \textbf{3.} Our results demonstrate large, statistically significant improvements in fairness across all datasets, tasks, and fairness sensitive variables: \textit{Race}, \textit{Age}, \textit{Sex}, and \textit{Marital status}.

\section{Methods}

\subsection{Design decisions for the fairness-enhancing, Mixed Effects Deep Learning framework, Fair MEDL}
\label{sec:fair_MEDL}
We constructed a framework that enables the development of predictive learning models that are simultaneously fair across subpopulations and account for clustered, non-\textit{i.i.d} data. To construct such a framework, we begin with a base architecture that can account for the clustering and then add compatible fairness enhancing subnetworks and modify the loss functions to simultaneously achieve both aims. There are several base MEDL architectures to choose from, such as: MeNet \cite{xiong2019}, LMMNN \cite{simchoni2021}, and ARMED \cite{nguyen2023ARMED}. Of these, the ARMED approach has been shown to have superior computational efficiency compared to other MEDL methods, improve performance on in-distribution data, enhance generalization for out-of-distribution (o.o.d) data, increase data interpretability, and mitigate confounding. For these reasons, we construct a base MEDL neural network inspired by \cite{nguyen2023ARMED}.

\subsection{The base MEDL architecture}
\begin{figure*}[htbp]
\centering
\includegraphics[width=0.7\linewidth]{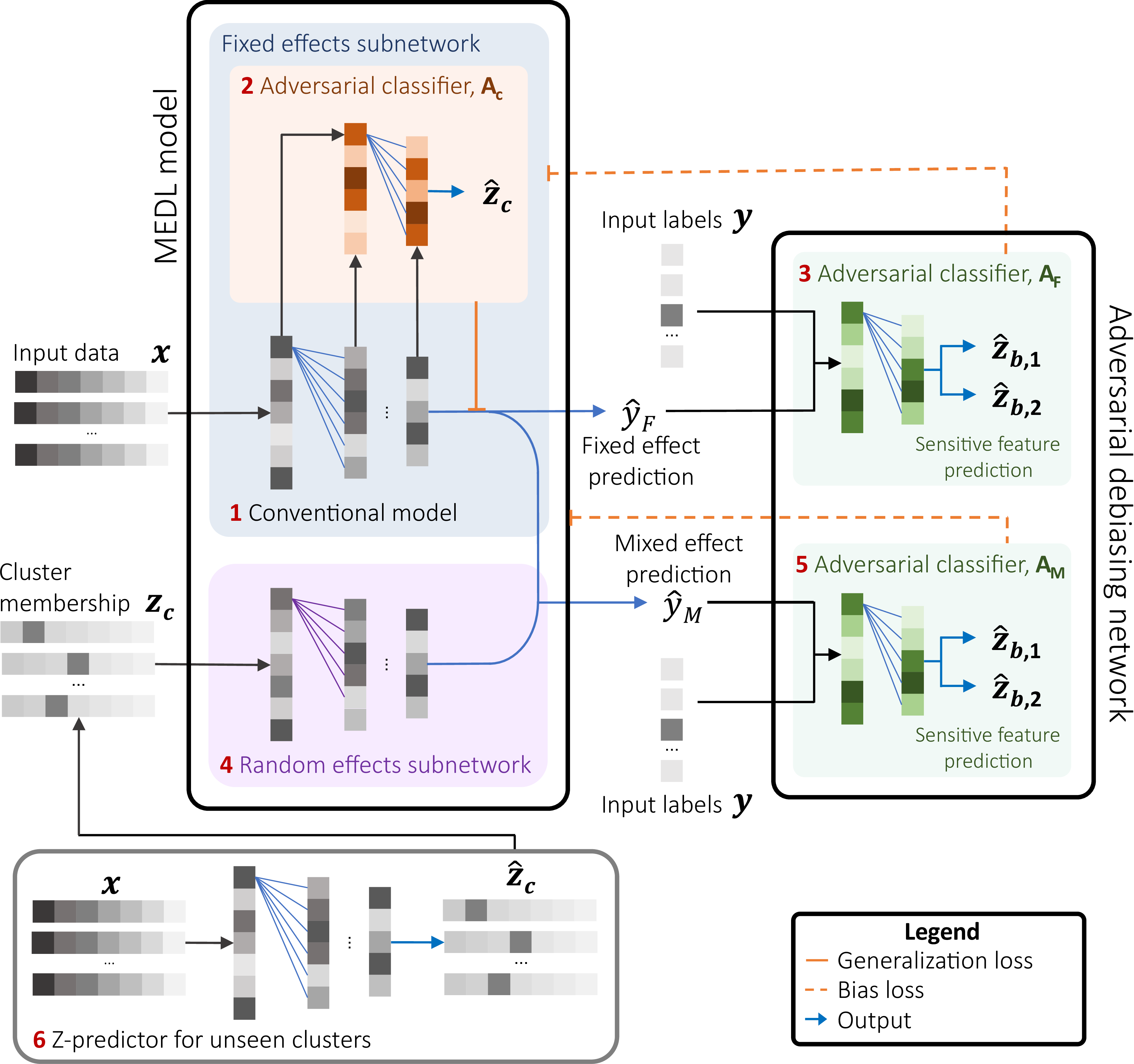}
\caption{Fairness enhancing MEDL framework with adversarial debiasing subnetworks for both the fixed, $A_F$, and mixed, $A_M$, effects.}
\label{fig:Fair_MEDL_Model}
\end{figure*}
We construct a base MEDL network for estimating the random effects as well as the fixed and mixed effects predictions. This feedforward neural network is shown in Fig. \ref{fig:Fair_MEDL_Model}, \textit{MEDL model}. This network has several components: 1) a core predictive conventional model ("1" in the figure) which predicts the target, $y$,  2) an adversarial classifier, $A_c$,  (Fig. \ref{fig:Fair_MEDL_Model}, subnetwork "2" which encourages the core conventional model to learn features \textit{not} predictive of the sample's associated cluster, thereby teaching the network to output a cluster-invariant fixed effects (FE) prediction, $\hat{y}_F$, 3) a random effects (RE) \textit{Bayesian} subnetwork (Fig. \ref{fig:Fair_MEDL_Model}, subnetwork "4" that learns to predict cluster-specific slopes and intercepts, denoted as RE, and 4) a mixing function that combines the FE and RE estimates into the cluster-specific mixed effect (ME) prediction, $\hat{y}_M$. For our base MEDL model, we also separately train a cluster membership ($z$) predictor "6" so that we can use our base MEDL network to predict on o.o.d data from clusters unseen during training (Fig. \ref{fig:Fair_MEDL_Model}, $z$-predictor subnetwork).

The objective function for our MEDL model is:
\begin{equation}
\begin{split}
&L_M(y, \hat{y}_M) + \lambda_F L_F(y, \hat{y}_F) - \lambda_g L_{CCE}(Z, \hat{Z}) \\
&+ \lambda_K D_{KL}(q(U) \parallel p(U))
\end{split}
\end{equation}
where $y$, $\hat{y}_M$, and $\hat{y}_F$ signify the ground truth label, mixed effects prediction, and fixed effects prediction, $L_M$ and $L_F$ are data fidelity constraints for the mixed and fixed effects predictions, respectively. $Z$ and $\hat{Z}$ are the cluster ID and its prediction, and $L_{CCE}$ denotes the categorical cross-entropy loss of the Adversarial Classifier (2) predicting the cluster. The term $D_{KL}(q(U) \parallel p(U))$ is the KL-divergence between the Bayesian neural network submodel's approximation, $q(U)$, and the prior posterior, $p(U)$. The hyperparameters $\lambda_F$, $\lambda_g$, and $\lambda_K$ are tuned using hyperparameter optimization as detailed in section \ref{sec:DA-frames}. Next, we describe how to incorporate fairness constraints into our base MEDL model and generalize it to regression tasks, capabilities lacking in prior works. 

\subsection{Fairness metrics: definitions and adaptation }
\label{sec:metrics}
At this point, the model may be able to handle mixed effects, but it does not encourage fairness. To address this, we must first define what we mean by fairness and then adapt the metrics for our needs. There are multiple measures of fairness, and in this work, we select three fairness metrics: equalized odds, demographic parity, and counterfactual fairness. These metrics comprehensively span a variety of aspects of fairness and, with some innovation, can be efficiently calculated and understood, for both classification and regression. \textit{Equalized odds} focuses on prediction accuracy, ensuring that error rates (false positives and false negatives) are similar across groups defined by sensitive attributes. This metric emphasizes minimizing bias in misclassification based on fairness-sensitive attributes. In contrast, \textit{demographic parity} measures the extent to which predictions are independent of sensitive attributes, with the aim of equalizing the probability of positive outcomes across groups. While equalized odds addresses fairness by balancing errors, demographic parity promotes fairness by balancing outcome distributions across groups. On the other hand, \textit{counterfactual fairness} provides a causal perspective, characterizing the extent to which predictions remain fair even when considering hypothetical (counterfactual) changes to the fairness attribute, while leaving all other covariates unchanged. These metrics are interconnected, as highlighted in recent literature, which emphasizes their complementary nature and common fairness context\cite{anthis2024, rosenblatt2023}. They can be implemented in neural networks or used to quantify the effects on fairness of newly proposed DL frameworks.  With some innovation, they can be incorporated into an in-process method to handle sensitive variables with more than two categories (e.g., race)\cite{yang2023adversarial}. Moreover, as we demonstrate, we can adapt these methods for both classification and regression. In the following subsections, we define the concepts behind these metrics, explain them for simple and complex classification tasks, and adapt them for regression.
\subsubsection{Equalized odds}
\label{sec: DA-DL}
Equalized odds\cite{hardt2016eqodds} states that fair predictions should have the same predictive power for all subgroups. Formally, in binary classification, for all pairs of sub-populations $s_1$ and $s_2$:
\begin{align}
    &P(\hat{Y}=1 | Y=1, S=s_1) = P(\hat{Y}=1 | Y=1, S=s_2)  \nonumber\\
    &P(\hat{Y}=0 | Y=0, S=s_1) = P(\hat{Y}=0 | Y=0, S=s_2)
\end{align}

where \( \hat{Y} \) is the outcome predicted by the model, and $Y$ is the real outcome. For each possible outcome, the probability of the model predicting the outcome should be the same regardless of which category (e.g., \textit{Race}) the samples are from.

In classification tasks, this metric is assessed by measuring the standard deviation of the true positive rate (TPR) and the false positive rate (FPR) across categories within each sensitive attribute, with maximal fairness achieved when the standard deviation is 0. For regression, the standard deviation of the mean squared error (MSE) value is used, rather than TPR and FPR. \\

\subsubsection{Demographic parity}
Demographic parity \cite{pessach2022review} is a fairness criterion for machine learning models that measures the extent to which a model's predictions are independent of a sensitive attribute (e.g., \textit{Race}, \textit{Sex}). Perfect demographic parity is achieved if the probability of receiving a positive prediction is the same across different groups defined by the sensitive attribute. Mathematically, when the number of such groups is 2, $S$ is the sensitive attribute, and $\hat{Y}$ is the predicted outcome (e.g., 0 or 1), demographic parity is defined as:
\begin{equation}
    \label{eq:demographic_parity_2cat}
	DP_{\text{class, 2 groups}} = \left| p(\hat{Y} = 1 \mid S = 1) - p(\hat{Y} = 1 \mid S \neq 1) \right| \nonumber
\end{equation}
where $DP$ is a value between 0.0 and 1.0. A lower value indicates better fairness, and perfect (ideal) DP=0.0. In general, when there are $|\mathcal{S}|$ groups comprising the fairness sensitive attribute (e.g., $|\mathcal{S}|$ races), demographic parity can be defined as the mean of all possible pairwise differences of  subpopulations in the data:
\begin{align}
    \label{eq:demographic_parity}
	&DP_{class} = \nonumber \\
    &\frac{1}{C(|\mathcal{S}|,2)} \sum_{s=1}^{|\mathcal{S}|-1} \sum_{s' = s+1}^{|\mathcal{S}|} \left| p(\hat{Y} = 1 \mid S = s) - p(\hat{Y} = 1 \mid S = s') \right|
\end{align}
In this equation:
\begin{itemize}
    \item \( \mathcal{S} \) represents the set of all subpopulations, $s_1,...,s_{|\mathcal{S}|}$.
    \item \( s \) is a specific subpopulation within \( \mathcal{S} \).
    \item \( |\mathcal{S}| \) denotes the cardinality of the set \( \mathcal{S} \), which is the number of different subpopulations (e.g., \textit{Races}) considered.
    \item \( C(|\mathcal{S}|,2)\) is the binomial coefficient which is the number of distinct pairs of subpopulations that can be chosen from \( \mathcal{S} \).

\end{itemize}
Equation \ref{eq:demographic_parity} can be extended to regression targets by changing the probability of the discrete outcomes to expected values of predicted outcomes:
\begin{align}
	&DP_{reg} = \nonumber \\
    & \frac{1}{C(|\mathcal{S}|,2)} \sum_{s=1}^{|\mathcal{S}|-1} \sum_{s' = s+1}^{|\mathcal{S}|} \left| E[\hat{Y} \mid S = s] - E[\hat{Y} \mid S = s'] \right|
\end{align}

where $E(\hat{Y})$ is the expected prediction of the continuous outcomes.
\subsubsection{Counterfactual fairness} 
Counterfactual fairness \cite{kusner2017} measures a model's fairness as the degree to which the model's predictions remain the same when we intervene by changing a sensitive attribute (e.g., \textit{Race}) to a counterfactual value, while keeping all other covariates the same. Formally, a predictive model is said to be counterfactually fair if, for all samples $x\in \mathcal{X}$, $\forall s\in \mathcal{S}, s' \in \mathcal{S} \setminus s, \text{and } \forall y$, the following condition holds:
\begin{equation}
        p(\hat{Y}=y \mid S = s, X = x) = p(\hat{Y}=y \mid S = s', X = x)
\end{equation}
where:
\begin{itemize}
    \item \( \mathcal{S} \) represents the set of all subpopulations.
    \item \( S \) is the subpoluation of the sample.
    \item \( X \) contains all other characteristics of the sample.
    \item \( s, s' \) are specific subpopulations within \( \mathcal{S} \). While \( (s, x) \) represents the measured data point,  \( (s', x) \) is a sample in a counterfactual world where sample  \(x\) is assigned an alternative  \(s'\)
    \item \( \hat{Y} \) is the outcome predicted by the model.
\end{itemize}
This condition asserts that the prediction \( \hat{Y} \) should be invariant to changes in the subpopulation membership \( S \), provided that all other measured covariates \( X \) remain constant. For a classification task, counterfactual fairness can be expressed quantitatively as:

\begin{align}
    \label{eq:CF_cat}
	&CF_{class} = \frac{1}{N\times (|\mathcal{S}|-1)} \sum_{n=1}^{N}  \sum_{s' \in \mathcal{S} \setminus s_n} \nonumber \\
	&\left| p(\hat{Y} = 1 \mid S = s_n, X = x_n)  - p(\hat{Y} = 1 \mid S = s'_n, X = x_n) \right|
\end{align}
where:
\begin{itemize}
    \item N is the total number of samples (observations)
    \item \( (x_n, s_n)\) is the \(n^{th}\) observation
    \item \( (x_n, s'_n)\) is a counterfactual of \(n^{th}\) observation
\end{itemize}
By changing the probability of the given discrete outcomes to expected values of predicted outcomes, equation \ref{eq:CF_cat} can be extended to regression targets:
\begin{align}
        &CF_{reg} = \frac{1}{N\times (|\mathcal{S}|-1)} \sum_{n=1}^{N}  \sum_{s' \in \mathcal{S} \setminus s_n} \nonumber \\ 
        &\left| E[\hat{Y} \mid S = s_n, X = x_n] - E[\hat{Y} \mid S = s'_n, X = x_n] \right|
\end{align}

\subsection{Making the fixed effects prediction fair} \label{sec:DA-frames}
The MEDL network introduced in section \ref{sec:fair_MEDL} is not yet fair. To promote fairness, we must implement both architectural changes and loss function modifications. We choose to implement equalized odds fairness directly in our framework. This metric is implemented because of its practical applicability and alignment with real-world fairness needs. (1) Equalized odds directly addresses both false positives and false negatives by ensuring that the false positive rate (FPR) and false negative rate (FNR) are equal across all demographic groups. (2) It aligns with real-world causal structure by disentangling the effects of the sensitive attribute on both the prediction and the true outcome. (3) It can be implemented through interpretable architectural changes and loss function modifications that explicitly encode fairness constraints directly into the DL model optimization. Additionally, implementing equalized odds not only encourages that form of fairness but can also promote counterfactual fairness and demographic parity. By reducing the dependency of the predicted outcome on the sensitive attribute, equalized odds reduces the likelihood that changing the sensitive attribute (in a counterfactual world) will result in a different prediction, minimizing the impact of the sensitive attribute. Additionally, promoting equalized odds also ensures conditional fairness, which often has the side effect of improving demographic parity. In practice, models trained to meet equalized odds will often make more balanced positive predictions across groups.

In this section, we describe how to make the fixed effects, $\hat{y}_F$, prediction fair, while in the subsequent section, we describe how to make the random effects prediction fair, thereby enabling the overall mixed effects prediction, $\hat{y}_M$, to be fair. Our implementation is based on equalized odds; however, we evaluate the effects of the implementation, using all three fairness metrics in the applications (section \ref{sec:applications}). We begin by forming a subset of the base MEDL network in Fig \ref{fig:Fair_MEDL_Model}, that is a domain adversarial (DA) model. This simpler model includes subnetworks 1 and 2, while subnetwork 4 is left out. See Fig. \ref{fig:Fair_MEDL_Model}). This facilitates rapid experimentation and obtaining greater insight into fairness enhancement. For the classification tasks, we build the DA model with the loss function:
\begin{align}
    \mathcal{L}_{\text{DA}}(\theta) =  
    \lambda_y \mathcal{L}_{BCE}(y_F, \hat{y}_F) - \lambda_z \mathcal{L}_{CCE}(Z, \hat{Z})
\end{align}
where the first and second terms are the binary cross-entropy and categorical cross-entropy losses for the FE classifier and the cluster adversary, $A_C$, respectively (Fig \ref{fig:Fair_MEDL_Model} network "2"). The hyperparameters $\lambda_y$ and $\lambda_z$ are determined via the Bayesian Optimization Hyperband (BOHB) search method \cite{falkner2018bohb}. For regression tasks, the first term is replaced with MSE loss. 

We make the domain adversarial (DA) model fair by constructing  a fairness-promoting adversarial debiasing subnetwork $A_F$ (Fig. \ref{fig:Fair_MEDL_Model}, subnetwork "3"). The $A_F$ subnetwork aims to predict the sensitive attribute of a sample (e.g., \textit{Race}) using the fixed effects subnetwork target prediction, $\hat{y}_F$,  and the true target $y$. If the error (i.e., difference) between $\hat{y}_F$ and $y$ is predictive of the group within the fairness sensitive attribute, for example, if the model makes more errors on a particular \textit{Race}, then it is unfair, and the FE network is penalized accordingly. In this way, $A_F$ encourages the FE network to learn weights that produce both accurate and fair predictions across subpopulations.

We denote the overall model, which combines the domain adversary components for fixed effects and the fairness-promoting adversarial subnetwork $A_F$, as the \textit{fair(ADB) DA-NNet} model. This model has the following loss objective:
\begin{align}
    L_{\text{fair DA adv. deb.}}(\theta) = & 
    \lambda_y L_{BCE}(y, \hat{y}_F) - \lambda_z L_{CCE}(Z_c, \hat{Z}_c) \nonumber\\
    &- \lambda_{FE} L_{CCE_{FE}}(S, \hat{S}).
\end{align}
The hyperparameter ($\lambda_{FE}$) for the fixed effects, adversarial debiasing subnetwork allows  tuning of fairness through hyperparameter optimization, which we perform via BOHB search \cite{falkner2018bohb}. 

There is an alternative approach to implement equalized odds fairness, which entails the introduction of a new loss function, known as the absolute correlation loss (ACL) \cite{beutel2019correlationRE}. Inspired by this approach, in lieu of adding an adversary debiasing network for the FE, we add a loss term which promotes fairness. For the new ACL loss term, a Pearson correlation coefficient is computed between the predictor output and values of the sensitive variable. Then, the correlation loss term using this coefficient can be added for each sensitive feature, incentivizing learning that reduces overall correlation, according to the following overall loss function:
\begin{align} 
	\label{eq:DA-ACL-loss}
    \mathcal{L}_{\text{fair DA ACL}}(\theta) &=  
	\lambda_y \mathcal{L}_{BCE}(y_F, \hat{y}_F) - \lambda_z \mathcal{L}_{CCE}(Z, \hat{Z}) \nonumber \\
	&+ \lambda_s (|\mathcal{L}_{corr}(\hat{y}_F, S)|
\end{align}
for sensitive features $S$  with an added correlation loss weight hyperparameter ($\lambda_s$). We denote this model as the \textit{Fair(ACL) DA-NNet} model, and we use it to explore the potential of this alternative approach.

\subsection{Making the random effects prediction fair}
Our proposed full framework, Fair(ADB) MEDL, requires an adversarial debiasing subnetwork, $A_F$, to make the fixed effect predictions fair, and an additional adversarial debiasing subnetwork, $A_M$, to make the mixed effects predictions fair. These are subnetworks 3 and 5 in Fig. \ref{fig:Fair_MEDL_Model}). In addition to these architectural changes, they require new loss terms. The overall loss function is:
\begin{align}
	\mathcal{L}_{\text{fairMEDL}}(\theta) &=  
    \mathcal{L}_M(y, \hat{y}_M) + \lambda_F \mathcal{L}_F(y, \hat{y}_F) - \lambda_g \mathcal{L}_{CCE}(Z, \hat{Z}) \nonumber\\
	&+ \lambda_K D_{KL}(q(U) \parallel p(U)) \nonumber\\
    &- \lambda_{FE} \mathcal{L}_{CCE_{FE}}(S, \hat{S}) - \lambda_{ME} \mathcal{L}_{CCE_{ME}}(S, \hat{S})
\end{align}
Here, $S$ and $\hat{S}$ correspond to the sensitive features and their predictions within the debiasing adversarial networks. The weights $\lambda_{FE}$ and $\lambda_{ME}$ modulate the loss of the adversarial debiasing subnetworks. These hyperparameters can also be efficiently tuned using BOHB \cite{falkner2018bohb}.

\begin{table}[htbp]
    \centering
    \footnotesize
    \caption{Adult dataset demographics subset}
    \label{tab:Adults_demographic_subset}
    \resizebox{\columnwidth}{!}{
    \begin{tabular}{lrrr}
        \toprule
        \textbf{Feature} & \textbf{Category} & \textbf{Count} & \textbf{Percentage} \\
        \midrule
        \textbf{Sex} & & & \\
        & Male & 21781 & 66.94\% \\
        & Female & 10757 & 33.06\% \\
        \midrule
        \textbf{Race} & & & \\
        & White & 27794 & 85.42\% \\
        & Black & 3123 & 9.6\% \\
        & Asian-Pacific-Islander & 1039 & 3.19\% \\
        & American-Indian-Eskimo & 311 & 0.96\% \\
        \midrule
        \textbf{Age Group} & & & \\
        & $<=30$ & 10558 & 32.45\% \\
        & $>30$ and $<=45$ & 12355 & 37.97\% \\
        & $>45$ & 9625 & 29.58\% \\
        \midrule
        \textbf{Occupation} & & & \\
        & Craft-repair & 4096 & 12.59\% \\
        & Exec-managerial & 4065 & 12.49\% \\
        & Sales & 3648 & 11.21\% \\
        & Transport-moving & 1596 & 4.91\% \\
        & Farming-fishing & 993 & 3.05\% \\
        \midrule
        \textbf{Income} & & & \\
        & $<=50k$ & 24707 & 75.93\% \\
        & $>50k$ & 7831 & 24.07\% \\
        \bottomrule
    \end{tabular}}
\raggedright
\scriptsize
{\textit{For complete demographics details, please refer to the supplemental Table S1}}
\end{table}

\subsection{Experiments}
\label{sec:applications}
We tested our models on three datasets: Adult, IPUMS, and Heritage Health, spanning different sectors (finance, healthcare) and tasks (classification and regression). All three datasets are publicly available and widely used for research purposes. They have been anonymized to protect individual privacy. The \textit{Adult dataset} \cite{misc_adult_2} derived from US Census data, contains 14 demographic and employment attributes. The classification target is to predict whether an individual earns over \$50,000 per year. The random effect, \textit{occupation}, captures a sample of 15 of the possible occupations and induces clustered sample correlation, as samples from the same occupation have similar attributes. Fairness-sensitive variables are \textit{Sex} (male, female), \textit{Age} bracket ($<$30, 31-45, $>$45), \textit{Race} (black, white, American Indian/Alaska Native, Asian, other), and \textit{Marital status} (6 categories). After data cleaning, the dataset contains 23,002 samples (individuals).  Table \ref{tab:Adults_demographic_subset} gives an overview of the Adult dataset demographics (Full demographics in Supplemental Table S1).

\begin{table}[htbp]
    \centering
	\footnotesize
	\caption{Heritage Health dataset demographics subset}
    \label{tab:HH_demographic_subset}
	\resizebox{\columnwidth}{!}{
		\begin{tabular}{lrrr}
		\toprule
            \textbf{Feature} & \textbf{Category} & \textbf{Count} & \textbf{Percentage} \\
            \midrule
            \textbf{DSFSclaim} & & Mean & Std \\
            & & 0.51 & 0.53 \\
            \midrule
            \textbf{Age} & & & \\
            & 0-29 & 8368 & 21.85\% \\
            & 30-49 & 8874 & 23.17\% \\
            & 50-69 & 9199 & 24.02\% \\
            \midrule
            \textbf{Sex} & & & \\
            & F & 17723 & 46.28\% \\
            & M & 13923 & 36.35\% \\
            \midrule
            \textbf{Specialty} & & & \\
            & Anesthesiology & 2293 & 5.99\% \\
            & Diagnostic Imaging & 16996 & 44.38\% \\
            & Emergency & 7977 & 20.83\% \\
            & General Practice & 22992 & 60.03\% \\
		\midrule
		\textbf{PrimaryConditionGroup} & & & \\
		& HEART2 & 2035 & 5.31\% \\
  		& HEART4 & 1955 & 5.1\% \\
		& NEUMENT & 10697 & 27.93\% \\
		& RENAL1 & 28 & 0.07\% \\
  		& SEIZURE & 1205 & 3.15\% \\
    		& TRAUMA & 4701 & 12.27\% \\
		\midrule
          \textbf{Top providers ID} & & &\\
              & 7053364 & 9311 & 12.25\% \\
            & 1076052 & 6584 & 8.66\% \\
            & 321261 & 3497 & 4.60\% \\
            & 4107701 & 2877 & 3.78\% \\
            \midrule
		\textbf{DaysInHospital} & Range& Mean & Std \\
		&0-15 & 0.46 & 1.62 \\
		\bottomrule
	\end{tabular}}
 \raggedright
\scriptsize
{\textit{For complete demographics details, see supplemental Tables S4, S5, S6, and S7}}
\end{table}

The \textit{IPUMS dataset} \cite{ipums_usa} was sourced from the 2006 American Community Survey census. This dataset includes records from 1,572,387 US residents, covers 14 socio-demographic attributes, and specifies whether an individual's income is $>$\$50,000. The samples are clustered by occupation (the random effect) with 15 different categories. Fairness-sensitive variables mirror those in the Adult dataset (\textit{Sex, Age, Race, Marital status}), but \textit{Race} is expanded to include additional classes (e.g., Chinese, Japanese, and mixed). Supplemental Table S2 highlights key demographic attributes of the IPUMS dataset (Full demographics in Supplemental Table S3). This dataset was used to test the \textit{scalability} of our proposed fairness-enhancing MEDL framework to a dataset \textgreater48x larger than the Adult dataset.

To test how well the proposed framework would generalize to data from a different sector (healthcare rather than finance) and regression rather than classification, we used the\textit{ Heritage Health dataset} \cite{heritage_health_2011}. This dataset contains healthcare insurance claims for patients affiliated with the California-based Heritage Provider Network. These records were aggregated into a yearly summary of all claims per patient, comprising 19 features (e.g., age, number of claims, place of service, clinical specialty). These summary records cover 76,013 patients. The target is the number of days spent in the hospital in the following year. The sensitive variables are \textit{Age} bracket (0-29, 30-49, 50-69, 70-79, 80+) and \textit{Sex} (male, female, NA), while the primary healthcare provider (2708 total providers) is the random effect. Table \ref{tab:HH_demographic_subset} presents the key demographic aspects of the Heritage Health dataset (Full demographics in Supplemental Tables S4-S7).

\subsection{Data partitioning and performance evaluation}
For model training, each dataset was partitioned into \textit{seen} and \textit{unseen} subsets based on the random effect. In the Adult dataset, the six highest-frequency occupations (71\% of samples) constituted the seen data. For IPUMS, the top five occupations (70\%) were used as seen cluster data, while for Heritage Health, the top 50 providers (50\%) were used. Data from the seen clusters were used for model training, and the remaining data in each dataset were held-out as unseen clusters to test model generalizability. For each dataset, seen data were further partitioned via 10-fold cross-validation, with 8 folds to train a model, one fold for validation, and one fold for testing. This testing data were used for computing the seen cluster performance metrics in the following \textit{Results} section. The test metrics are comprehensive, as they are compiled from the samples across all 10 folds. Depending on the task (e.g., classification or regression) model performance was measured using balanced accuracy or MSE, respectively, while fairness was evaluated using all three measures: equalized odds, demographic parity, and counterfactual fairness, as defined in section \ref{sec:metrics}.

To assess what the models learned, we evaluated feature importance by calculating the gradient magnitude with respect to each input feature \cite{dimopoulos1995,olden2004}. To assess the model's effectiveness in mitigating confounding factors, we introduced probes that were correlated with the target and cluster but not correlated with any of the original dataset covariates, so the probes did not have domain (financial or health) information. Our hypothesis is that the baseline deep learning models would rely heavily on the probes to predict outcomes, thereby producing type I errors, whereas our proposed model would identify the confounding probes and significantly reduce their influence, mitigating the confounding.

\section{Results}
\subsection{Income prediction using the Adult dataset}
\begin{figure*}[htbp]
    \centering
    \includegraphics[width=\linewidth]{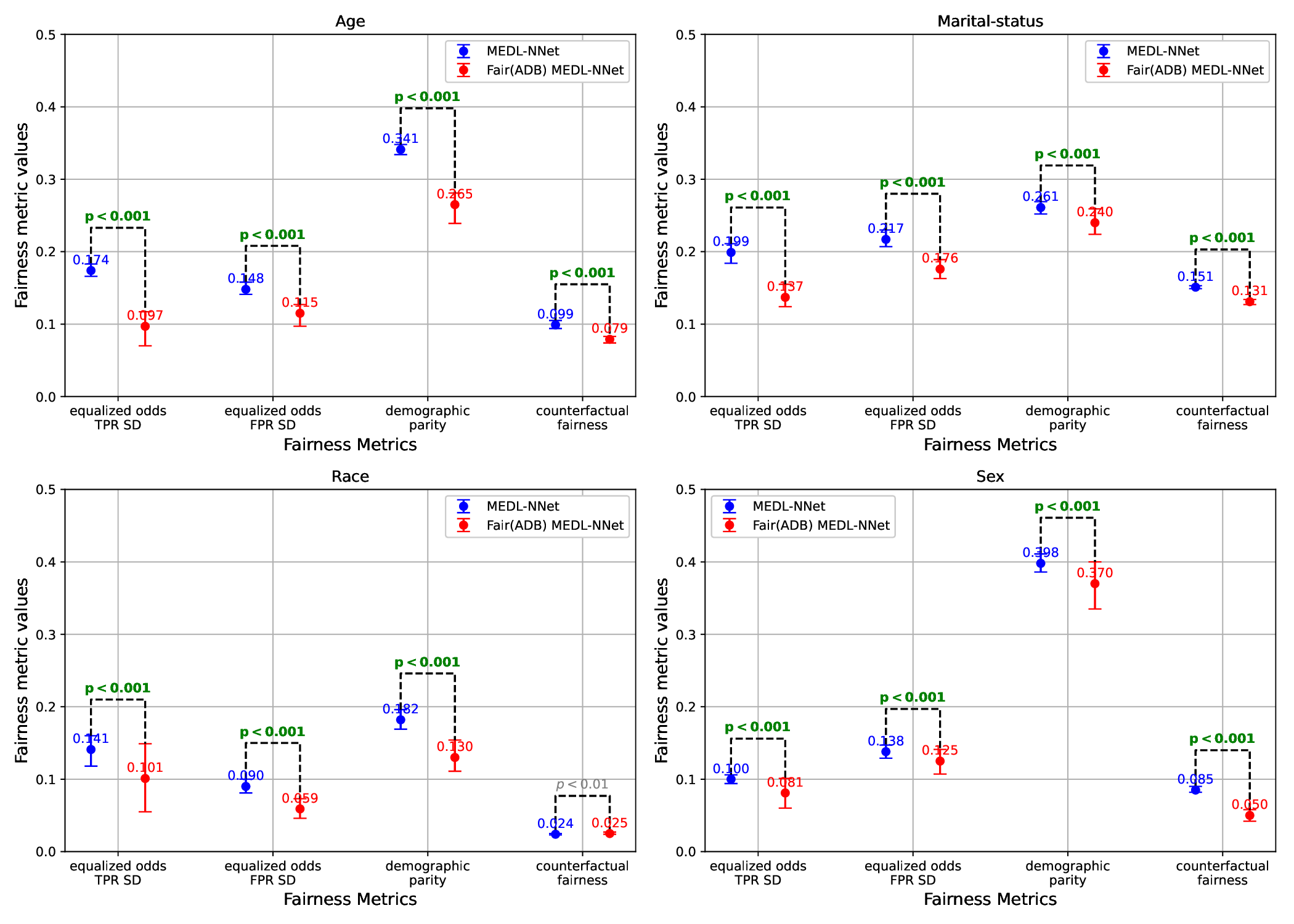}
    \captionsetup{justification=centering}
    \caption{Adult dataset, Fair(ADB) MEDL-NNet improves fairness metrics across different sensitive variables}
    \label{fig:Adult_result}
    \begin{minipage}{\linewidth}
        \centering
        \scriptsize
        TPR SD : True positive rate standard deviation, FPR SD: False positive rate standard deviation. \\ \textbf{Bold green} font color indicates moderate-large practical effect, Grey font color indicates negligible practical impact.
    \end{minipage}
\end{figure*}

In this section, we describe the results of the models performing a \textit{classification} task in predicting whether or not an individual's income level is above \$50,000, using the Adult census dataset, and how the proposed fairness-enhancing MEDL framework improves fairness for the task across all of the dataset's fairness-sensitive variables, for all three fairness metrics. Figure \ref{fig:Adult_result} presents the results across four fairness-sensitive attributes: \textit{Age}, \textit{Sex}, \textit{Race}, and \textit{Marital status}. We observe that the proposed framework, Fair(ADB) MEDL-NNet, improved fairness for all four attributes while maintaining high prediction accuracy. For the sensitive attributes \textit{Age}, \textit{Marital status}, and \textit{Sex}, substantial improvements are observed in \textbf{all} fairness metrics. Specifically, the TPR standard deviation (SD) for \textit{Age} improved by 44.3\%, with a desirable reduction from 0.174 to 0.097 (See Supplemental Tables S8 and Table S9 for numerical results). Similarly, the counterfactual fairness for \textit{Sex} improved by 41.2\%, reducing from 0.085 to 0.05. Meanwhile, for \textit{marital status}, the TPR and FPR improved by 31.2\% and 18.9\%, respectively, while the counterfactual fairness improved by 25.2\%. 
For \textit{Race}, three of the fairness metrics showed substantial improvements, with TPR and FPR improving by 28.4\% and 34.4\%, respectively, and demographic parity improving by 28.6\%. Only counterfactual fairness remained essentially unchanged, with a negligible increase from 0.024 to 0.025. Meanwhile, in terms of model accuracy, as shown in supplemental Table S10, we observe that, while the original, unfair MEDL model had an AUROC and accuracy of 0.890 and 0.813, respectively, the proposed model has nearly the same AUROC and accuracy of 0.882 and 0.811, respectively. Thus, while we see large improvements in fairness, model performance remains largely stable. 

The alternative approach, absolute correlation loss (ACL), demonstrates inconsistent performance in improving fairness, as shown in supplemental Tables  S16 and S17, where it improves 18 out of 32 and 16 out of 32 fairness values, respectively. In contrast, the results in supplemental Table S8, S9, S11, and S12 highlight that the proposed adversarial debiasing approach, Fair(ADB), achieves consistently strong improvements across all considered fairness metrics. Specifically, Fair(ADB) improves 46 out of 48 fairness values for the Adult dataset and 45 out of 48 for the IPUMS dataset. This consistency is further supported by results from the Heritage Health dataset (supplemental Table S15), where Fair(ADB) improves all 6 out of 6 fairness values.

\begin{figure*}[h]
    \centering
    \includegraphics[width=0.9\linewidth]{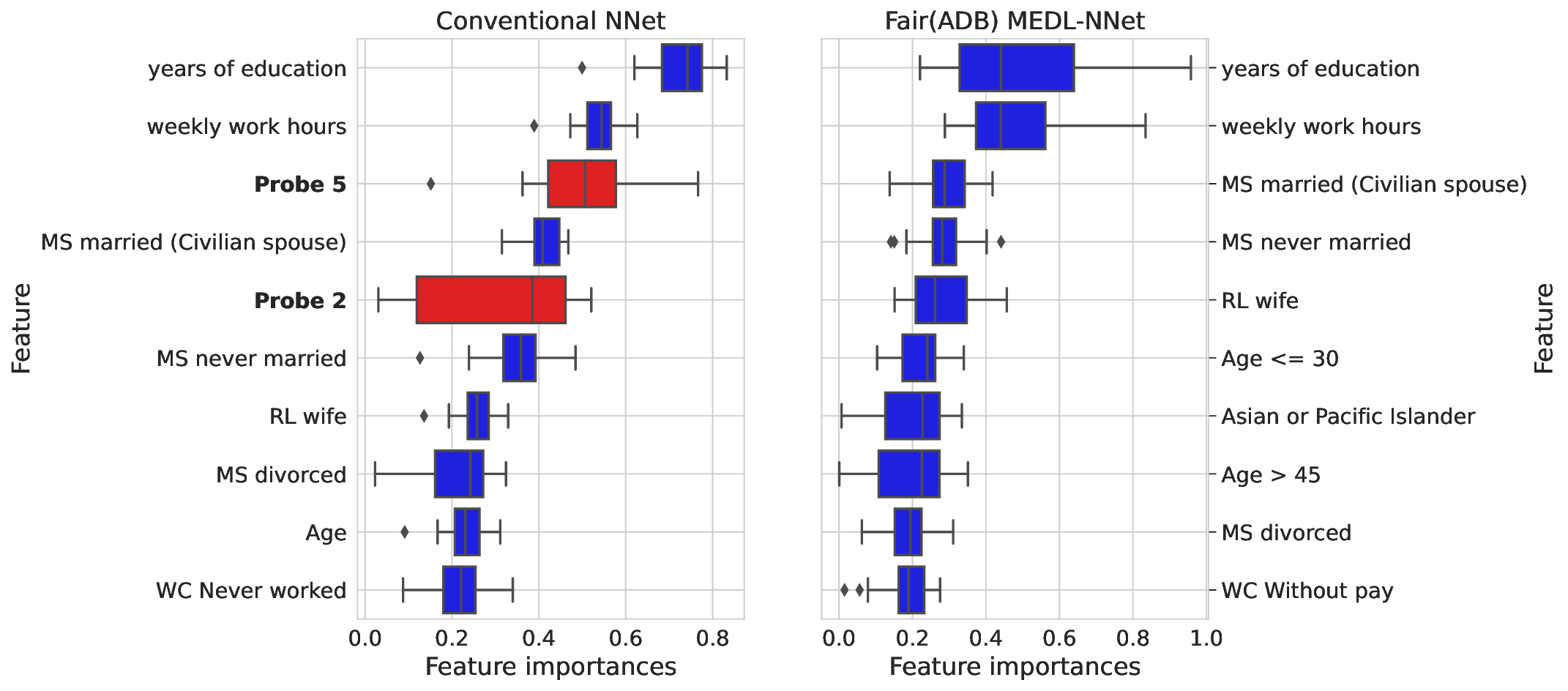}
    \captionsetup{justification=centering}
    \caption{Adult dataset, features importances with probes}
    \label{fig:Adult_probe}
    \begin{minipage}{\linewidth}
        \centering 
        \scriptsize
        MS : Marital Status, RL: Relationship, WC: Working class
    \end{minipage}
\end{figure*}
The demonstration of the capacity to reduce Type I and II errors is a key benefit of MEDL. Figure \ref{fig:Adult_probe} highlights the ability of the fair MEDL framework to effectively de-weight the probes.
We observe that, While the conventional NNet ranks two probes as the $3^{rd}$ and $5^{th}$ most important features, the Fair(ADB) MEDL-NNet model does not rank any probe among its top 10 most important features. Upon further inspection, we note that features that it identified as important by the proposed model, look quite reasonable, with \textit{years of education} and \textit{weekly work hours} identified as the most influential features for predicting income level. This underscores our proposed model's ability to not only to identify the correct key features but also to avoid confounding effects—features that may appear important due to spurious correlations with both the target and site information. 
\subsection{Income prediction using the IPUMS data}

\begin{figure*}[htbp]
    \centering
    \includegraphics[width=\linewidth]{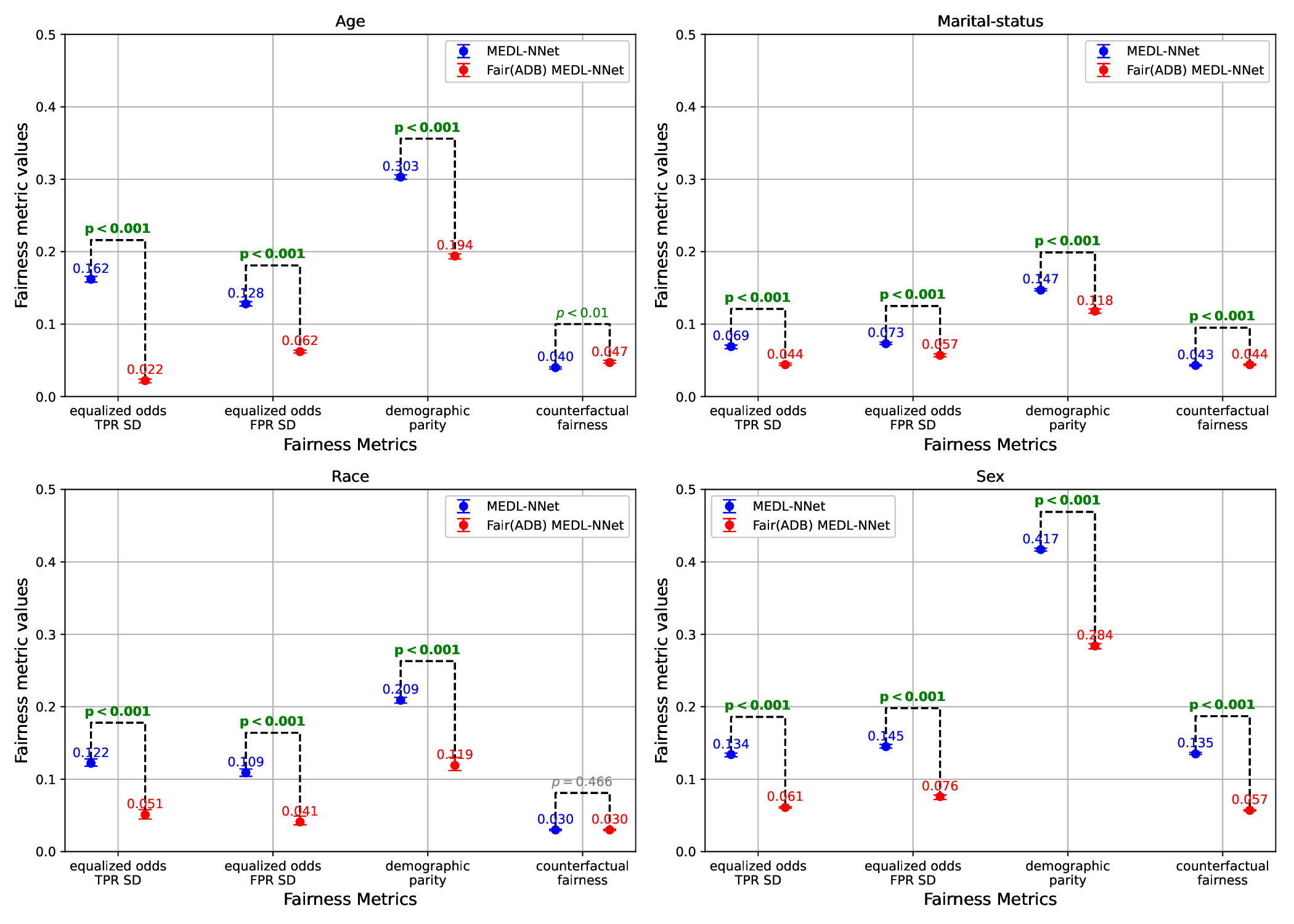}
    \captionsetup{justification=centering}
    \caption{IPUMS dataset, Fair(ADB) MEDL-NNet improves fairness metrics across different sensitive variables}
    \label{fig:IPUMS_result}
    \begin{minipage}{\linewidth}
        \centering
        \scriptsize
        TPR SD : True positive rate standard deviation, FPR SD: False positive rate standard deviation.\\ \textbf{Bold green} font color indicates moderate-large practical effect, \textit{Italics green} low practical effect, Grey font color indicates negligible practical impact.
    \end{minipage}
\end{figure*}

Regarding the equalized odds metric, the results in Figure \ref{fig:IPUMS_result} show that the proposed fairness-enhancing MEDL framework has good scalability to much larger data. In particular, the IPUMS dataset is over 48x larger than the Adult dataset and we still observe a substantial improvement in prediction fairness. For the proposed Fair(ADB) MEDL-NNet, for the full mixed effects prediction, the standard deviations of TPR and FPR for each fairness-sensitive variable show a marked reduction compared to the standard MEDL framework model, the MEDL-NNet. In this larger dataset, we observe this improvement for \textit{every} fairness-sensitive variable. For example, our results on the test folds from the clusters (occupations) seen during training, show the TPR standard deviation for \textit{Age} decreased from 0.162 to  0.022 in fair(ADB) MEDL predictions, a 86.4\% improvement in fairness. Similarly, for the \textit{Race}, the TPR standard deviation was reduced from 0.122 to 0.051 in fair(ADB) MEDL predictions, a 58.2\% improvement. Further, for the sensitive variable \textit{Sex}, we observe a 54.5\% enhancement in fairness, with TPR standard deviations dropping from 0.134 in MEDL-NNet to 0.0061 for the Fair(ADB) MEDL-NNet prediction.  A similar improvement in fairness for all variables is observed for the fixed effects (FE) predictions from the Fair(ADB) MEDL-NNet, (top 2  model rows, Table S11). Moreover, these improvements are not limited to the occupations seen (in-distribution data) during training.  On data from occupations unseen during training (OOD data), we find similar enhancements (bottom half, Table S11). Across \textit{Age} brackets, we observe an impressive 62.9\% improvement in TPR fairness (from 0.272 in the MEDL-NNet FE to 0.101 in the Fair(ADB) MEDL-NNet FE). Across categories of \textit{Race}, fairness improves by approximately 43.4\%, with standard deviations dropping from 0.145 in the MEDL-NNet FE to 0.082 in the Fair(ADB) MEDL-NNet FE. Similarly, excellent results are observed for the Fair MEDL ME predictions (see supplemental Table S11) where the full (ME models) are denoted as the MEDL-NNet model and the proposed Fair(ADB) MEDL-NNet model.

Importantly, these reductions in standard deviations for both TPR and FPR across various sensitive attributes are \textit{statistically significant}, with p-values consistently less than 0.001 when compared against the baseline MEDL FE and ME models. Meanwhile, balanced accuracy exhibits only minor fluctuations as shown in supplemental Table S13. We find this marginal trade-off reasonable given the substantial improvements to fairness. Supplemental Table S12 presents the results for demographic parity and counterfactual fairness. The Fair(ADB) MEDL framework shows significant improvements for \textit{all} sensitive features across both the FE and ME predictions regarding demographic parity. In terms of counterfactual fairness, we observe a notable enhancement for \textit{Sex}, with a remarkable 57.8\% improvement, reducing from 0.135 to 0.057. Similar improvements are evident for \textit{Race} and \textit{Marital status}. However, we observe a slight decrease in fairness for \textit{Age}. 
\begin{figure*}[h]
    \centering
    \includegraphics[width=\linewidth]{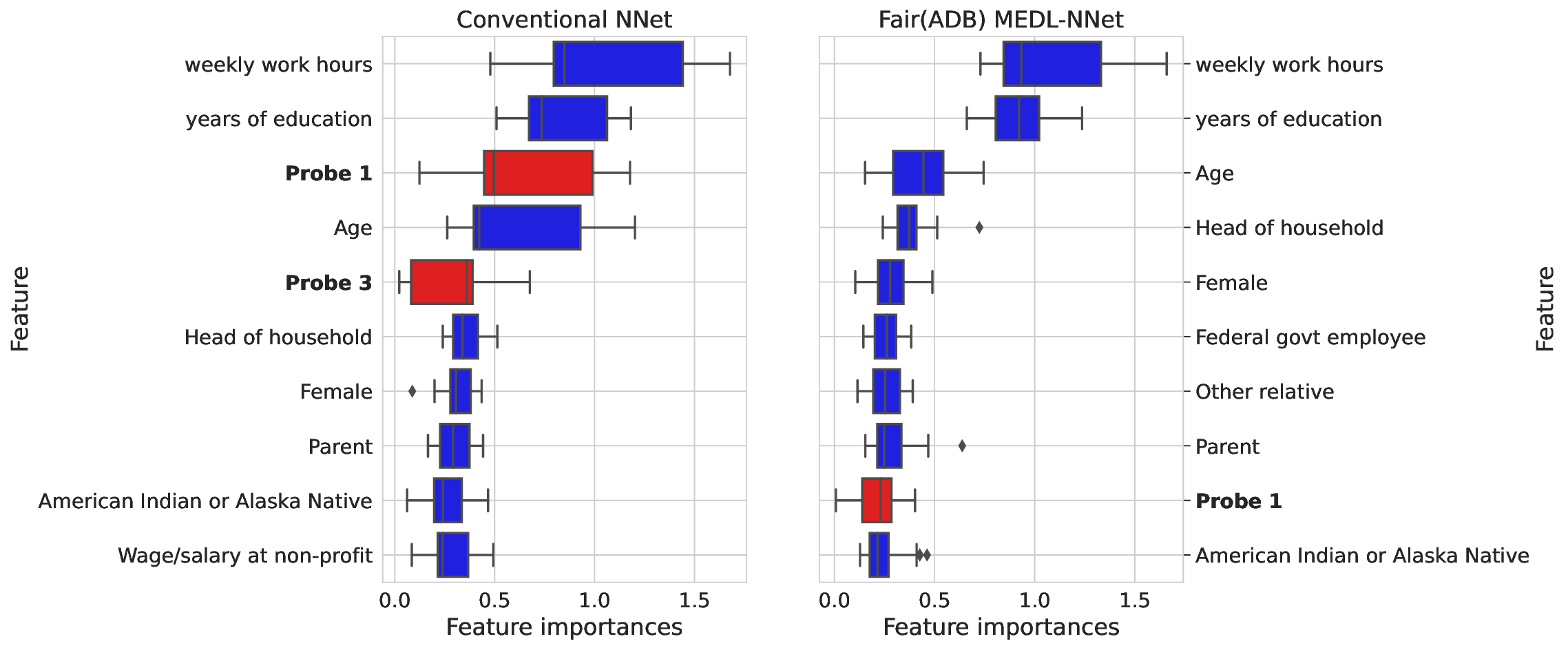}
    \captionsetup{justification=centering}
    \caption{IPUMS dataset, features importances with probes}
    \label{fig:IPUMS_probe}
\end{figure*}
In addition, the Fair MEDL framework continues to demonstrate exceptional ability to identify and de-weight confounding probes, indicative of its ability to mitigate confounds. As shown in Figure \ref{fig:IPUMS_probe}, the conventional NNet ranks two probes among its top five most important features. In contrast, the proposed Fair(ADB) MEDL-NNet reduces the influence of these probes, with only one probe remaining, which was ranked $9^{th}$ overall. Here, we observe Probe 1 being de-weighted significantly, dropping from $3^{rd}$ place in the conventional NNet to $9^{th}$ in the proposed fair model. Additionally the features the proposed model identified as most important, appear sound, with \textit{weekly work hours} and  \textit{years of education} identified as the top 2 features.

\subsection{Future hospital visit prediction using the Heritage Health dataset}
\begin{figure*}[h]
    \centering
    \includegraphics[width=0.95\linewidth]{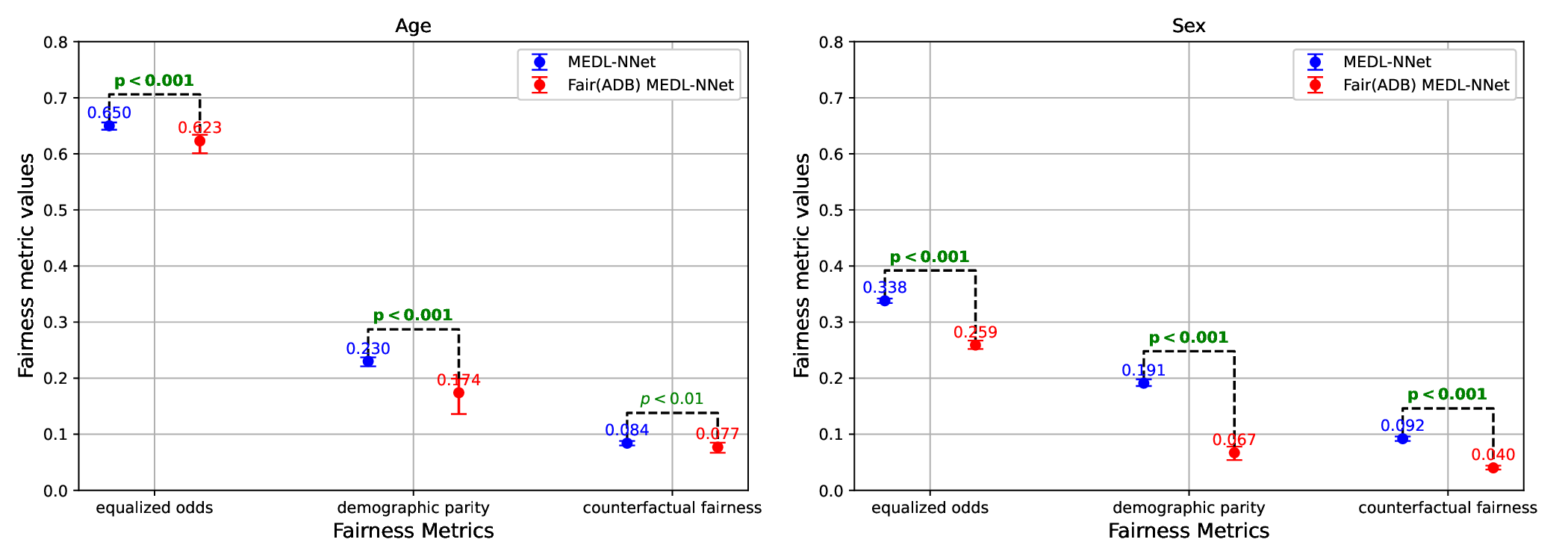}
    \caption{Heritage Health dataset, fairness metrics across different sensitive variables: age and sex}
    \label{fig:HH_result}
    \begin{minipage}{\linewidth}
        \centering
        \scriptsize
         \textbf{Bold green} font color indicates moderate-large practical effect, \textit{Italics green} font indicates low practical effect.
    \end{minipage}
\end{figure*}
The results on the Heritage Health dataset demonstrate the capacity of the proposed framework to improve fairness for a \textit{regression} task across all of the dataset's fairness-sensitive variables (\textit{Age} and \textit{Sex}), for all three fairness metrics, as shown in Figure \ref{fig:HH_result} and Table S15. Note that, for regression, these metrics are measured somewhat differently compared to classification (Section \ref{sec:metrics}).

All fairness metrics improved, with the most significant enhancement observed for counterfactual fairness for \textit{Sex}, with a substantial 56.5\% reduction from 0.092 for the MEDL-NNet to 0.040 for the Fair(ADB) MEDL-NNet. This is particularly noteworthy, given that counterfactual fairness was the one metric with varying outcomes in the IPUMS dataset. This underscores that fairness improvements are also data-dependent, similar to many other challenges. Nevertheless, the proposed Fair(ADB) MEDL-NNet consistently outperforms the traditional MEDL neural network across datasets in nearly all attributes and measures. Moreover, supplemental Table S14 shows negligible changes in the prediction performance.

\begin{figure*}[htbp]
    \centering
    \includegraphics[width=\linewidth]{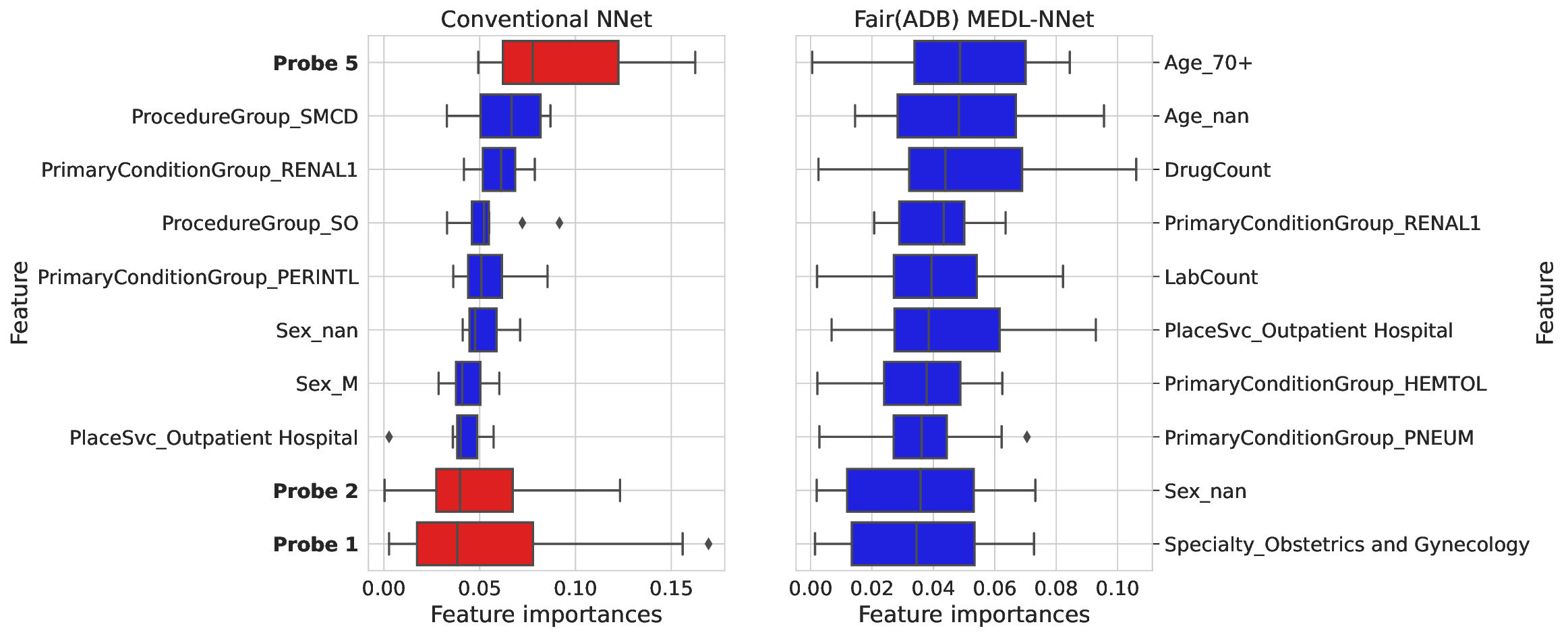}
    \caption{Heritage health dataset, features importances with probes}
    \label{fig:HH_probe}
\end{figure*}

The proposed framework also demonstrated good abilities to mitigate confounds for regression, complementing the classification results above. Figure \ref{fig:HH_probe} illustrates  the Fair(ADB) MEDL-NNet's ability to de-weight confounding probe features. While the conventional NNet ranks three of the five probes among its top 10 most important features, the Fair(ADB) MEDL NNet successfully eliminates all probes from its top rankings.
\subsection{The full model outperforms the ablated model}
A domain adversarial ablated model, Fair(ADB) DA-NNet, was constructed which captures fixed effects through a FE subnetwork but lacks the RE subnetwork (Fig. \ref{fig:Fair_MEDL_Model}, network "4")  for the modeling of random effects. In our head-to-head comparisons, the full proposed framework achieved greater performance, with 1.3\% higher balanced accuracy (on average) in predicting income level on the Adult dataset, as shown in Table S18, right column, underscoring the benefit of retaining all of the full framework's components. Moreover, the full framework achieves greater fairness in 7 out of 8 fairness values (Table S18, TPR and FPR standard deviation columns) compared to the Fair(ADB) DA-NNet model. Additionally, compared to non-fair models, a trade-off in accuracy often occurs, as fairness enhancement introduces an additional fairness loss term which can compete with the data fidelity term, unless there are additional covariate combinations can which satisfy both. Notably, we observe that the average trade-off is more pronounced in the ablated framework with a 1.03\% absolute reduction in balanced accuracy (Fair(ADB) DA vs plain (unfair) DA model, Table S18) compared to the proposed framework with just a 0.25\% absolute reduction in accuracy (Fair(ADB) MEDL vs MEDL model, Tables S10 and S18.

\section{Discussion}
The proposed Fair(ADB) MEDL-NNet framework preserves the advantages of mixed effects deep learning, which include a better ability to understand both the data and the predictive model (\cite{nguyen2023ARMED}). For example, on the Adult dataset the framework demonstrated the unique capacity to distinguish between occupation-agnostic effects learned through the FE subnetwork, and occupation-specific effects learned through the RE subnetwork. The proposed unified framework preserves these benefits while substantially increasing fairness in the FE and ME predictions from the MEDL model. This unified approach enables the framework to provide insight into both the \textit{fair} cluster-specific effects and the \textit{fair} cluster-agnostic effects. That is, the proposed framework offers a deeper level of interpretability compared to conventional neural networks, which typically do not differentiate between these effects, and greater fairness compared to traditional MEDL frameworks, which are not intrinsically fair. The proposed fair MEDL framework meaningfully advances both fairness and interpretability, offering more nuanced insights into what the model has learned.

The proposed Fair(ADB) MEDL-NNet framework includes subnetworks that explicitly separate and quantify the fixed and random effects. This enables the framework to simultaneously identify and de-weight confounding probe features, and provides it with the capacity to reduce Type I and II errors. While the parallel subnetworks for capturing fixed and random effects increase overall model complexity and require additional hyperparameter tuning, we have found that this can be readily addressed using the BOHB search method \cite{falkner2018bohb}. Additionally, by using modern distributed computation frameworks for compute clusters, such as Ray Tune\cite{liaw2018tune}, the hyperparameter tuning can be  implemented and solved efficiently.

Recently, Yang et al. (\cite{yang2023adversarial}) enhanced fairness for a non-mixed effects network. Their method showed a consistent fairness improvement in predicting COVID infection. However, their approach demonstrated only a relatively modest increase in fairness of up to 5\% compared to the conventional unfair network. In addition, while the Yang et al. work is noteworthy for its suitability for infection prediction, the work was only demonstrated on a classification task. In contrast, in this work, the proposed fair MEDL framework consistently achieves much greater fairness enhancement across datasets from different sectors (finance, healthcare), and across both classification and regression tasks, with a substantial improvement in fairness ranging from  0.3\%-86.4\% for \textit{Age},  7.0\%-57.8\% for \textit{Sex}, 0.5\%-64.9\% for \textit{Race}, and 0.4-36.2\% for \textit{Marital status}. 

We suspect that such marked improvement arises because our framework has more capacity to learn, separates fixed from random effects, and because we have optimized the trade-off between fairness and accuracy by exploring a greater number of hyperparameter configurations.

Overall, the framework results  demonstrate substantial improvements in fairness, with valuable implications for critical applications. Income prediction from datasets such as Adult and IPUMS is illustrative of the potentially life-altering predictions that machine learning models are making increasingly frequently, as income prediction plays a central role in financial decisions, such as whether an applicant will be given a loan or approved for a line of credit. By applying our framework, such predictions can become significantly fairer, ensuring equitable treatment across key demographic groups. Similarly, predicting future hospital utilization for individuals in the Heritage Health dataset, plays a central role in the calculation of insurance premiums, typically a major household expense. Fairness in these predictions is crucial, as it directly impacts individuals' financial obligations for healthcare. In this way, our framework can contribute to more equitable decision making in crucial sectors that affect quality of life.

\section{Conclusion}
This work represents the first framework to simultaneously address the need to enhance fairness while mitigating confounding and ameliorating the bias from clustered, non-iid data. We demonstrate how the loss function and architectural contributions of adversarial debiasing are fully compatible with a mixed effects (MEDL) framework, promoting fairness for both its fixed and random effects predictions. We showed how the benefits of the MEDL, including (1) increased interpretability of the model and data, (2) high performance on in-distribution (seen) and out-of-distribution (unseen cluster) data, and (3) mitigation of confounds to reduce Type I and II errors, are \textit{retained}, while new benefits, including \textit{substantially increased fairness}, are attained. We additionally show how fairness through adversarial debiasing has performance benefits over other types of fairness methods, such as absolute correlation loss. We have demonstrated how the benefits of fairness are even greater using the full proposed framework compared to an ablated model, the Fair(ADB) DA-NNet model, which does not explicitly model the random effects. We also illustrated how fairness enhancement can be formulated for regression (not just classification), and how the proposed framework improves three fairness metrics: equalized odds, demographic parity, and counterfactual fairness.

The framework's benefits (increased fairness, high prediction accuracy, increased interpretability, and confound mitigation), are \textit{consistent} across all fairness sensitive variables (e.g. \textit{Sex}, \textit{Age}, \textit{Race}, \textit{Marital status}), and across three distinct datasets, including two census datasets focused on an income prediction classification task and a healthcare dataset with a regression prediction target of the number of days an individual will spend hospitalized in the subsequent year. As such targets are commonly used in life-changing loan applications and healthcare insurance decisions, they can have profound societal implications; thus, ensuring the fairness of predictive models that underpin decision making in these sectors is paramount. Our results strongly support the recommendation of the Fair (ADB) MEDL-NNet framework, as it allows stakeholders to reap the many salient benefits of MEDL while ensuring fairness for all individuals. We make our code available at \href{https://tinyurl.com/FairMEDL}{\texttt{tinyurl.com/FairMEDL}}

\section*{Acknowledgments}
This study was supported by NIH grant R01GM144486 and the Lyda Hill Foundation. Data used in this paper include the following: 1) the Adult ("Census Income") dataset, extracted by Barry Becker, and available on the UC Irvine Machine Learning Repository, 2) the Heritage Health dataset, which is available through the Heritage Health Prize competition archives, and was collected by the Heritage Provider Network of California, and 3) the IPUMS data, which was extracted from the 2006 American Community Survey Census, provided by the US Census Bureau.

\bibliography{fairness_medl}
\bibliographystyle{IEEEtran}
\vskip -3\baselineskip plus -1fil

\begin{IEEEbiography}[{\includegraphics[width=1in,height=1.25in,clip,keepaspectratio]{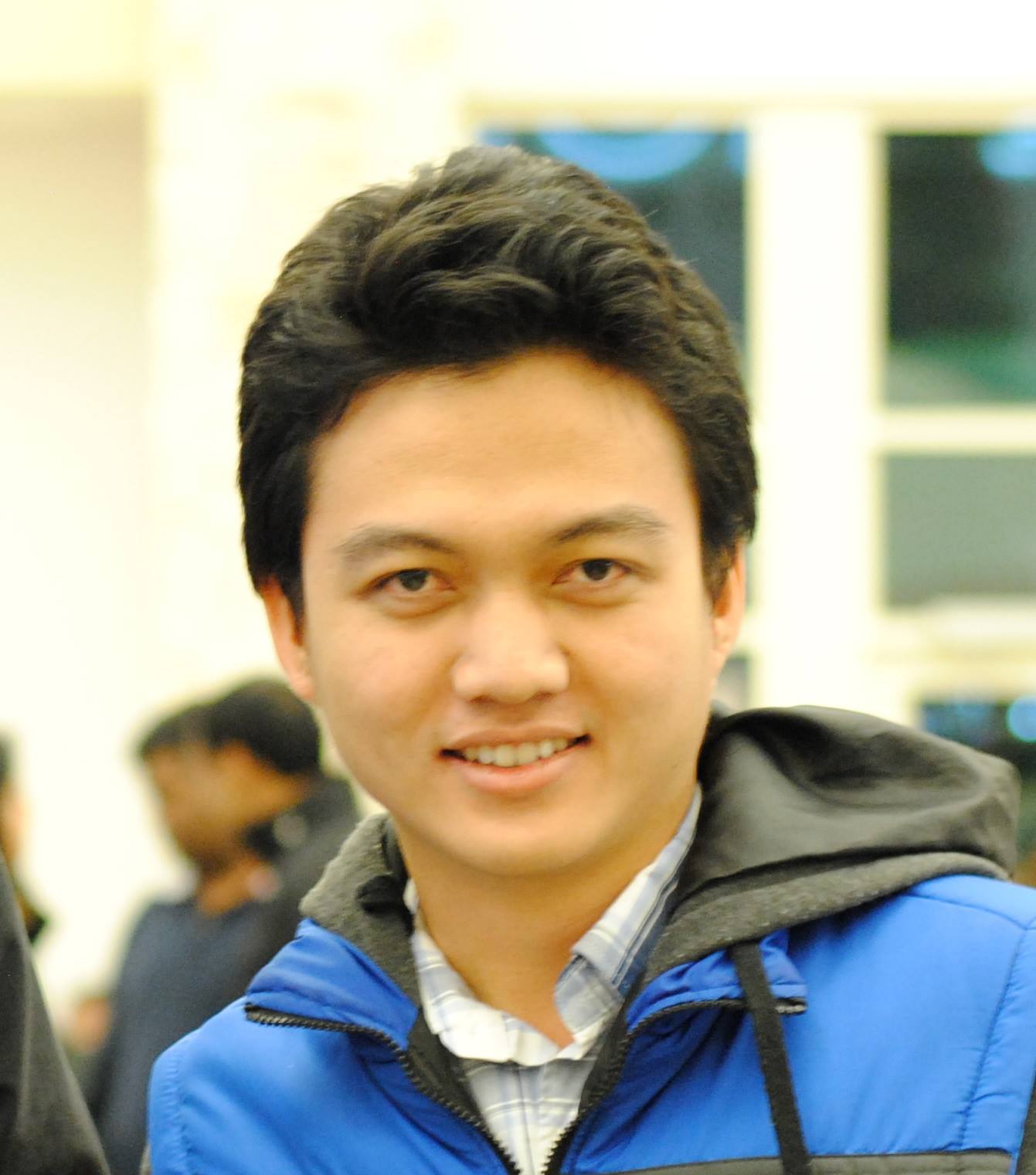}}]
{Son N. Nguyen}
earned his PhD in Electrical Engineering with a specialization in machine learning from the University of Texas at Arlington. He is currently a postdoctoral fellow focusing on developing novel methods to enhance fairness and causal inference in deep learning, as well as leveraging deep learning techniques for improved medical diagnosis.
\end{IEEEbiography}

\vskip -4\baselineskip plus -1fil

\begin{IEEEbiography}[{\includegraphics[width=1in,height=1.25in,clip,keepaspectratio]{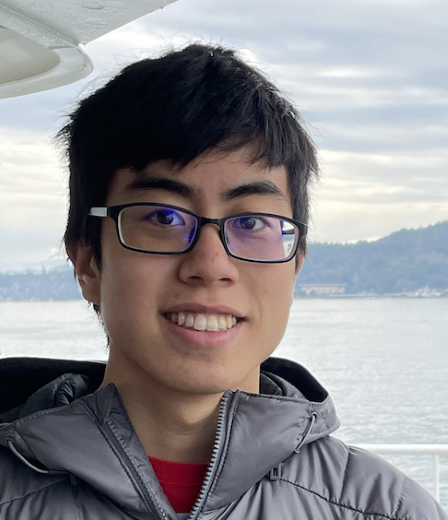}}]{Adam J. Wang} is a student at Harvard University, working towards a degree in Bioengineering with concentrations in Electrical Engineering and Computer Science. His research interests include  technological and biological systems with precision health applications, from high performance machine learning models to gene interactions at the single cell resolution.
\end{IEEEbiography}

\vskip -4\baselineskip plus -1fil

\begin{IEEEbiography}[{\includegraphics[width=1in,height=1.25in,clip,keepaspectratio]{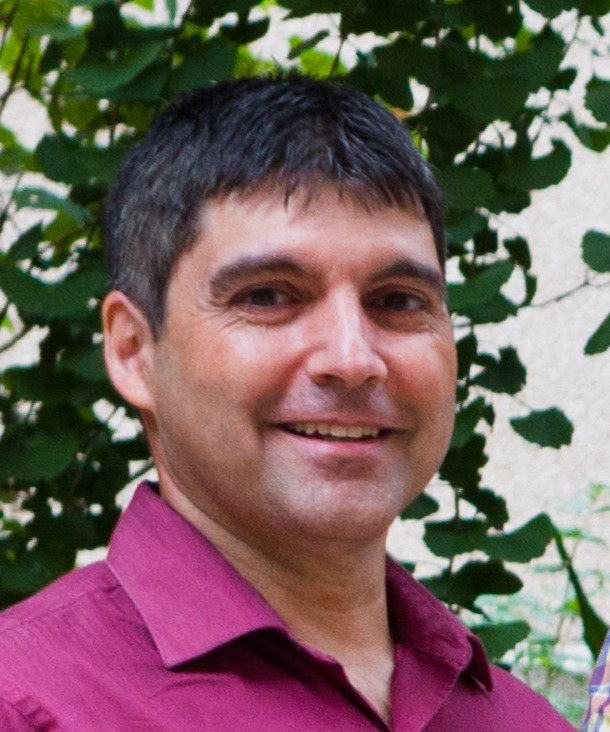}}]{Albert A. Montillo}
received his BS and MS from Rensselaer and PhD in Computer Science and Medical Imaging from the University of Pennsylvania. He is an Associate Professor in the Departments of Bioinformatics and BME at UT Southwestern. His Deep Learning for Precision Health Lab develops the theory of machine learning for improved interpretability and generalization especially for prognostics from multimodal, multiomic data of neurological disorders and improved disease mechanism insights. 
\end{IEEEbiography}

\end{document}